\documentclass{article}
\usepackage{amssymb}
\title{Stereotypical Reasoning: Logical Properties}
\author{Daniel Lehmann
\\Institute of Computer Science, Hebrew University, 
\\Jerusalem 91904, Israel. \\E-mail:~lehmann@cs.huji.ac.il}
\date{24 October 1997.}

\newtheorem{theorem}{Theorem}
\newtheorem{corollary}{Corollary}

\newtheorem{example}{Example}

\newcommand{\blackslug}{\mbox{\hskip 1pt \vrule width 4pt height 8pt 
depth 1.5pt \hskip 1pt}}
\newcommand{\QED}{\quad\blackslug\lower 8.5pt\null\par\noindent}
\newcommand{\proof}{\par\penalty-100\vskip .5 pt\noindent{\bf Proof\/: }}

\newcommand{\ru}{\rule[-0.4mm]{.1mm}{3mm}}
\newcommand{\nni}{\ru\hspace{-3.5pt}}

\newcommand{\NI}{\mbox{$\: \nni\sim$}}
\newcommand{\cC}{\mbox{${\cal C}$}}
\newcommand{\Cn}{\mbox{${\cal C}n$}}
\newcommand{\eqdef}{\stackrel{\rm def}{=}}

\begin{document}
\maketitle

\begin{abstract}
Stereotypical reasoning assumes that the situation at hand is 
one of a kind and that it enjoys the properties generally associated
with that kind of situation.
It is one of the most basic forms of nonmonotonic reasoning.
A formal model for stereotypical reasoning is proposed and the logical
properties of this form of reasoning are studied.
Stereotypical reasoning is shown to be cumulative under weak assumptions.
\end{abstract}

\section{Introduction}
Common sense reasoning in AI requires drawing inferences in a bolder,
more adventurous way, than mathematical reasoning.
Many different formalisms that implement some form of bold reasoning
have been proposed, implemented, used to build artificial systems.
Almost no work has been done comparing those formalisms with the way
natural intelligence deals with those tasks.
Minsky~\cite{MinskyWin:75,MinskyHaug:81} 
represents probably one of the only efforts to model
reasoning performed by natural intelligence.

During the last decades, philosophers, linguists, sociologists
have revolutionized the way we understand the human mind.
Putnam~\cite{Putnam:Mind} has criticized the classical philosophical 
assumptions about
meaning, and claimed that stereotypes are a necessary component of
the meaning of terms.
Rosch~\cite{Rosch:73,Rosch:75a,Rosch:81} 
has put in evidence the essential function of categorization in
achieving intelligence
and the intricate ways in which we use it.
Categorization is the process in which we relate a specific object or
situation to the kind we shall think it a member of.
She showed that many of our categories have prototypes, i.e., best
examples.
Lakoff~\cite{Lakoff:Women} resumes and expands much of this line of research.

The purpose of this work is to begin the study of inferencing in
a mind that uses categories as described above.
A model of the simplest kind of inferencing using stereotypes will
be given and the formal properties of the inferencing process will
be studied.
The formal properties of inferencing that are of interest have
been singled out by~\cite{Gabbay:85,KLMAI:89,Mak:Handbook}.
\section{Stereotypical Reasoning}
In this work, stereotypical reasoning is used to denote what is
probably the simplest form of natural nonmonotonic reasoning.
The present use of the word stereotype is very closely related
to Putnam's stereotypes. He claimed that stereotypes are a necessary
part of the meaning of words denoting a {\em natural kind}.
Here, stereotypes are assumed not only for natural kinds but for
any state of information.
This could be understood as the assumption that stereotypes are
part of the meaning of any sentence, but the philosophical aspects
of this assumption are not discussed in this paper.
The point of this paper is the study of how stereotypes are used
in the inferencing process, and the formal properties of the inferencing
resulting from the use of stereotypes. The use of stereotypes
has not been discussed by Putnam.

What is called here stereotypical reasoning is very closely related
to the use of what Rosch calls prototypical categories.
Prototypical alludes, though, to a richer structure than stereotypical
and this is the reason the latter term has been preferred.
In ordinary parlance stereotypes are considered to be typically
wildly inaccurate and an impediment to intelligent thinking.
This reputation should not hide the fact that the use of stereotypes
is a fundamental tool, probably the central tool, in achieving intelligence.
Hence, the importance of its study.
Nevertheless, the negative connotation attached to the word {\em stereotype}
should remind us we are studying a limited form of reasoning,
certainly not capable of exhibiting all forms of intelligence. 

Here is an example of what I will call stereotypical reasoning.
The choice of the {\em tiger} stereotype follows Putnam.
If Benjamin tells you that during his trip in India, hiking in the jungle, 
he saw a tiger, you will assume he saw a large, frightening animal,
yellow with black stripes.
Note that not all tigers are such. Some tigers are small, dead, or albino.
You have been using the stereotype that says that tigers are big, dangerous
and yellow with black stripes. The use of this stereotype may be a mistake:
the end of the story may reveal this was an albino tiger,
but, typically, the use of the stereotype is precisely what enables
efficient communication, since Benjamin knows you have this stereotype
(as he has) he assumes you will draw the corresponding conclusions and
he intends you to draw those conclusions.

This simple example already suggests a number of questions,
most of them will not be touched upon in this paper.
What is the nature, or the structure of the stereotype {\em tiger}?
Is it just a conjunction (or some other composition) of properties?
In this work, we shall assume that yes, a stereotype is a set of
possible states of affairs, but the proper treatment of prototypical
categories in general may need a more sophisticated structure.
Note, though, that we are not assuming (the classical view, attacked by Lakoff
and others) that categories are sets (of models), far from that.
We are only assuming that stereotypes are sets of models. 
In fact, the way stereotypes are used makes them function 
very much like the graded or
radial categories of Lakoff.

How are such stereotypes acquired? This is certainly both deeply rooted
in our physiology and a social process.

Why is this the {\em right} stereotype for tiger? Why is it any better
than some other? Here, an analysis of rationality and utility is certainly
needed.

What made you apply the {\em tiger} stereotype to the little story above
and not, for example, the {\em jungle} stereotype that says that, in the
jungle, everything is dark green, or the {\em India} stereotype, 
whatever this is
for you? In this paper this choice will be modeled by some distance
between the information at hand and the stereotype.
We shall not be able to explain why a specific distance is used.
Such an explanation would certainly be based on utility considerations.
\section{A formal model of stereotypical reasoning}
Informally, starting from information about the situation at hand,
one chooses the best stereotype to fit the information and uses both
the original information and the stereotypical information to draw conclusions.

Formally, we assume $W$ is the set of all possible states of affairs
(i.e., models or situations).
We assume a collection (not necessarily finite, but it may well be finite)
of stereotypes, \mbox{$S_{i}$}. Notice we use lower indexes to identify
the stereotypes. Each stereotype is a subset of $W$,
the set of situations in which the stereotype holds.
For example the {\em tiger} stereotype, $S_{tiger}$ could be the set of all
models in which tigers are frightening live animals, yellow with black stripes.

The user has some information, i.e., facts, about the situation at hand.
This information is modeled by a subset $F$ of $W$: the set of all situations
compatible with the information at hand.
On the basis of $F$, the reasoner picks up one of the stereotypes: $S^{F}$,
the stereotype most appropriate to $F$, in a way that will be discussed
later. 
Notice we use here an upper index to denote the stereotype that best fits
some information $F$.
The reasoner will then conclude that the actual state of affairs
is one of the members of the intersection \mbox{$F' \eqdef F \cap S^{F}$}.
The nonmonotonicity of the reasoning stems from this jump from $F$ to
the subset $F'$.
Clearly, we do expect the set $F'$ to be non-empty, assuming $F$ is non-empty,
since we want to avoid jumping to contradictory conclusions.
It will be the task of the function that defines the best stereotype
to pick a stereotype that has a non-empty intersection with the information
$F$ at hand.
In many cases the facts $F$ are given by a formula $\alpha$ that is known to
be true. In this case $F$ is the set of all models that satisfy $\alpha$.
We shall identify the formula $\alpha$ and the set of its models and write
$S^{\alpha}$ for the stereotype most appropriate for the sets of models
of $\alpha$. A formula $\beta$ is then nonmonotonically deduced from
$\alpha$ iff it is satisfied by all elements of $F'$, 
that is iff any model $m$ in the set $S^{\alpha}$ that satisfies  
$\alpha$ satisfies $\beta$: \mbox{$\alpha \NI \beta$} 
iff \mbox{$\forall m \in S^{\alpha}$},
\mbox{$m \models \alpha$} implies \mbox{$m \models \beta$}.

Syntactically, this may be described as taking for \mbox{$\cC(X)$},
the set of nonmonotonic consequences of a set $X$ of formulas,
the set \mbox{$\Cn(X, g(X))$}, 
of all formulas that logically follow from the set
\mbox{$X \cup g(X)$}, where \mbox{$g(X)$} is the set of formulas
that hold in all models of the stereotype that best fits $X$.

Our analysis of stereotypical reasoning will use the simplistic 
model just described, since it is good enough for the purpose of this
paper. If one thinks of first-order languages and models, one may want
to refine this model and associate a stereotype 
with each one of the objects of the structure: e.g., if our story refers
to two tigers about which one has different information, one will perhaps
use different stereotypes for each of the tigers: a mother tiger stereotype
and a pup tiger stereotype for example.

Before we analyze some consequences of this model, let us point out
some of its basic limitations.
It is assumed that the conclusions from facts $F$ are drawn by identifying
a {\em unique} stereotype most appropriate for $F$. One may ask whether
this should be the case.
Instead of picking up a single stereotype, perhaps one should consider the
set of all most appropriate stereotypes and use them all, i.e. their
intersection.
Indeed the results of next section would hold also in this more general
model, but the uniqueness assumption will be needed later.
Intuitively it seems to me that we do pick up a unique stereotype,
sometimes made up of different stereotypes, but that this composition is almost
never the simple juxtaposition, i.e., conjunction of stereotypes.
The main reason probably is that such conjunctions are very often empty
and we certainly want to avoid drawing inconsistent conclusions from consistent
facts.
Consider, for example, our tiger above.
We did not use both the {\em tiger} and the {\em jungle} stereotypes
because they clash about the color of the tiger: yellow and black vs. 
dark green. It may be the case that
a very smart reasoner will use a {\em tiger in the jungle} stereotype,
that implies the tiger is barely visible, but this stereotype,
though including, somehow, 
both the {\em tiger} and the {\em jungle} stereotypes,
cannot be reduced to their conjunction.
To avoid premature commitment to a theory of the formation of compound
stereotypes, this paper will just assume any set $F$ of facts 
is associated with a unique stereotype.
\section{First consequences of the model}
As general as it is, the model presented above has some important consequences
for the formal properties of the process of nonmonotonic deduction
it defines, i.e., of the consequence relation \NI.

First, since what is defined by facts $F$ is a set of models, $F'$,
the set of nonmonotonic consequences of $F$, i.e., the set of formulas
that hold in all elements of $F'$ is a logical theory, i.e., closed
under logical consequence. In other terms, the relation \NI\ satisfies
the rules of Right Weakening and And of~\cite{KLMAI:89}.
Secondly, since $F'$ is a subset of $F$, any formula that is logically
implied by $F$ holds in all elements of $F'$, or, the relation \NI\ satisfies
the Reflexivity of~\cite{KLMAI:89}.
Lastly, since the information at hand is represented, semantically, by a set
of models, the relation \NI\ satisfies Left Logical Equivalence.
\section{Further assumptions}
The main purpose of this work is to consider whether other, more sophisticated,
logical properties may be expected from stereotypical reasoning.
It is clear that, in the very general model described above, without
further assumptions, nothing more can be expected:
given any relation \NI\ satisfying Left Logical Equivalence, Reflexivity,
Right Weakening and And, one may define, for any formula $\alpha$,
the stereotype $S^{\alpha}$ to be the set of all models that satisfy all
the formulas $\beta$ such that \mbox{$\alpha$ \NI $\beta$}.
Since the relation \NI\ is reflexive, all models of $S^{\alpha}$ satisfy
$\alpha$ and, for any $\alpha$, \mbox{$S^{\alpha} \subseteq F^{\alpha}$}. 
Therefore \mbox{$F^{\alpha} \cap S^{\alpha} = S^{\alpha}$}
and the nonmonotonic consequence relation defined by the model is exactly
\NI.

Our goal is to find some additional, reasonable, assumptions about
the set of stereotypes or the way the best stereotype for a set $F$ is
chosen that will have interesting consequences on the nonmonotonic
logic defined.
In fact, the set of stereotypes and its structure does not seem to
play an important role here and we shall concentrate on the choice
of the best stereotype.

Notice that the mapping from a set $F$ to its best stereotype $S^{F}$
may be very wild. We do not expect, for example, that 
\mbox{$F' \subseteq F$} should imply \mbox{$S^{F'} \subseteq S^{F}$}.
It may well be the case that {\em robins} is the best stereotype for birds,
but the best stereotype for antarctic birds is, for lack perhaps of knowledge
of a better one, {\em vertebrates}. 

It is extremely helpful to consider the process of associating to the set
$F$ the stereotype $S^{F}$ as based on some notion of distance between
{\em information sets} and {\em stereotypes}: the best stereotype
for $F$ is the stereotype closest to $F$:
\begin{equation}
\label{eq:close}
d(F, S^{F}) \leq d(F, S) , {\rm \ for \ every \ stereotype \ } S.
\end{equation}
Notice that this notion of distance is a bit unusual, since it is defined 
only from information sets to stereotypes. We shall never use the notion
of the distance from a stereotype to an information set.
The assumption that our choice is based on some notion of a distance does
not limit the generality of our model, since one may always find a suitable
distance to fit any choice of best stereotype.
The interest of this assumption is that it suggests some natural
additional assumptions on the properties of this distance.
Those assumptions will be related to logical properties of the
nonmonotonic deduction.

Let us suppose $D$ is a partially ordered set (of distances)
and that there is a function $d$ that associates an element of $D$
\mbox{$d(F,S)$} with every set of models $F$ (in fact every set of models that
could appear as an information set would be enough) and every stereotype $S$.
A first assumption, already described above, is that this distance
always enables us to pick a unique best stereotype.
\begin{itemize}
\item {\em Assumption zero}: for any given information set $F$ there
exists a {\em unique} stereotype $S^{F}$ such that
\mbox{$d(F , S^{F}) \leq d(F , S)$} for any stereotype $S$.
\end{itemize}
A number of examples of models and choice functions will be described.
They may not be very intuitively appealing, but their purpose is 
to help the reader understand our definitions and prove the consistency
of the assumptions that shall be made below.
In all examples the set $D$ of distances is taken to be the set of integers,
eventually with $\infty$ added.
\begin{example}
\label{ex:1}
There is one stereotype only: $S_{0}$ and \mbox{$S_{0} = W$}.
The exact definition of the distance is irrelevant.
Assumption zero is satisfied trivially and, for any $F$, 
\mbox{$S^{F}= S_{0} = W$}.
Clearly, for any $F$, \mbox{$F' = F \cap S^{F} = F$}, and therefore
$F'$ is non-empty if $F$ is.
The non-monotonic logic defined happens to be monotonic and to
be the classical one: \mbox{$\alpha$ \NI $\beta$} iff
\mbox{$\alpha \vdash \beta$}.
\end{example}
\begin{example}
\label{ex:2}
Assume the set $W$ is finite.
Every set \mbox{$S \subseteq W$} is a stereotype and
\[
d(F, S) \: = \: \mid S - F \mid \: - \: \mid S \cap F \mid,
\]
where \mbox{$\mid A \mid$} indicates the cardinality of the set $A$.
Since \mbox{$d(F, F) = - \mid F \mid \leq d(F, S)$} for any $S$,
we see that, for any $F$, \mbox{$S^{F} = F$}, and therefore
assumption zero is satisfied, \mbox{$F' = F$} and the logic defined
is the classical one as in Example~\ref{ex:1}.
\end{example}
\begin{example}
\label{ex:3}
Assume the set $W$ is the set of natural numbers.
Stereotypes are singletons of $W$.
Distances are defined in the following way:
if \mbox{$n \in F$}, \mbox{$d(F, \{n\}) = n$} and
if \mbox{$n \not \in F$}, \mbox{$d(F, \{n\}) = \infty$}.
Clearly $S^{F}$ is the singleton that contains the minimal element of $F$,
\mbox{$\min(F)$} and assumption zero is satisfied.
Note also that \mbox{$F' = \min(F)$} is non-empty if $F$ is non-empty.
The model boils down to considering that world $m$ is more probable
than world $n$ iff \mbox{$m < n$}.
The logic defined results in, given a set of possibilities $F$, jumping to
the conclusion that the most probable one must obtain.
This provides a highly nonmonotonic consequence relation.
\end{example}
The next example presents a simple, but natural, family of models.
\begin{example}
\label{ex:4}
Assume $W$ is finite and the set of (non-empty) stereotypes 
\mbox{$S_{i}$}, \mbox{$i =0 , \ldots , k-1$} provides a
partition of $W$, i.e., \mbox{$\bigcup_{i \in k}S_{i} = W$}
and \mbox{$S_{i} \cap S_{j} = \emptyset$}, for any \mbox{$i \neq j$}.
Given a set $F$, we associate with it the stereotype $S_{j}$ which
{\em covers} $F$ best, i.e., for which the size of the set
\mbox{$S - F$} is minimal. In case this criterion does not define
a unique stereotype, choose the stereotype with smallest index.
Formally we may define the distance by:
\mbox{$d(F, S_{i}) = \mid S_{i} - F \mid + {i \over k}$}.
The consequence relation defined is nonmonotonic.
\end{example}
After these examples, let us consider interesting properties
of the distance $d$.
Since $F$ and $S$ are both sets of models (subsets of $W$) we may, without
loss of generality, assume that 
\mbox{$d(F , S) = e(F \cap S , S - F , F - S)$}.
Three additional assumptions concerning the way the function $e$ 
depends on each
of its three arguments are now natural.
\begin{itemize}
\item {\em Assumption one}: the function $e$ is anti-monotone in its first
argument. I mean that if \mbox{$A \subseteq A'$}, then
\mbox{$e(A' , B , C) \leq e(A , B , C)$}.
The relation $\subseteq$ is the subset relation.
This assumption is very natural: $d(F, S)$ measures the closeness
of $F$ and $S$: the more they have in common,
the closer they are. In most cases we expect that the best stereotype
for $F$ should be consistent with $F$, i.e., have an non-empty intersection
with $F$. If this is the case, our assumption is only slightly stronger: 
all other things being 
equal, the best stereotype for $F$ has the largest intersection with $F$.
The set \mbox{$F \cap S^{F}$} represents the nonmonotonic consequences
of $F$; we prefer weaker consequences, therefore we prefer to take the
set \mbox{$F \cap S^{F}$} as large as possible. 
\item  {\em Assumption two}: the function $e$ depends monotonically
on its second argument. Here I mean that if \mbox{$B \subseteq B'$}, then
\mbox{$e(A , B , C) \leq e(A , B' , C)$}.
The second argument, \mbox{$B = S - F$} measures the set of models
compatible with the stereotype but excluded by the information.
Notice that the stereotype may be vague, i.e., contain a large number
of elements: for example the {\em bird} stereotype may include birds of many
colors, and the information at hand may exclude a lot of those elements:
for example we may know the bird we are discussing is yellow.
The more such elements are excluded by the information at hand,
the less suitable is the stereotype: if too many such elements are excluded
a more specific stereotype may be more suitable.
In our example, a {\em yellow bird} stereotype, 
if there is one such stereotype, should be preferred.
\item  {\em Assumption three}: the function $e$ does not depend on its third
argument. 
It seems easy to justify that the function $e$ should depend monotonically
on its third argument, by an argument very similar to that used
for justifying assumption two.
It is perhaps a little less obviously natural that $e$ should not depend
at all on its third argument.
But, notice that the set \mbox{$F - S$} is a measure of the strength
of our nonmonotonic inference: the larger it is the more nonmonotonic
consequences we get in addition to the monotonic ones.
The argument just above is to the effect we should not get too many
such inferences, but
we are certainly interested in getting such nonmonotonic consequences,
and should not try to minimize them.
Our assumption is that how much nonmonotonicity we get should not 
be a criterion in choosing
the best stereotype.
\end{itemize} 

Assumptions one to three may be summarized by the following:
for any $F$, $F'$ and any stereotypes $S$, $S'$, if
\begin{equation}
\label{eq:onethree}
F' \cap S' \subseteq F \cap S
{\rm \ and \ } S - F \subseteq S' - F' {\rm \ then \ }
d(F, S) \leq d(F' , S').
\end{equation}
\proof
\[
d(F, S) = e(F \cap S, S - F, F - S) \leq
e(F' \cap S', S' - F', F' - S') = d(F', S')
\]
\QED
One may notice that
Equation~\ref{eq:onethree} implies that \mbox{$d(F, S) = d(F \cap S, S)$}.
Let us consider the examples above again.
In Example~\ref{ex:1}, we may define the distance $d$ to be constant,
for example \mbox{$d(F, S) = 0$}.
Equation~\ref{eq:onethree} is obviously satisfied.
In Example~\ref{ex:2} also, Equation~\ref{eq:onethree} is obviously satisfied.
In Example~\ref{ex:3}, let us check that
Equation~\ref{eq:onethree} is satisfied.
Since stereotypes are singletons, 
\mbox{$F' \cap S' \subseteq F \cap S$} implies that either
\mbox{$F' \cap S' = \emptyset$}, or \mbox{$S' = S = F' \cap S' = F \cap S$}.
In the first case \mbox{$ d(F' , S') = \infty$} and the result holds.
In the second case, if \mbox{$S = \{ n \}$}, 
\mbox{$d(F, S) = n = d(F', S')$}.
For Example~\ref{ex:4}, if \mbox{$F' \cap S' \subseteq F \cap S$} then,
either
\mbox{$F' \cap S' = \emptyset$}, or \mbox{$S' = S$}.
If \mbox{$S - F \subseteq S' - F'$}, then either \mbox{$S - F = \emptyset$}
or \mbox{$S' = S$}.
If \mbox{$S' = S = S_{i}$}, 
\[
d(F, S) \: = \: \mid S - F \mid \: + \: {i \over k} \: \leq \: 
\mid S' - F' \mid \: + \: {i \over k} \:
= \: d(F', S').
\]
If \mbox{$S' = S_{j} \neq S = S_{i}$}, \mbox{$F' \cap S' = \emptyset$} and 
\mbox{$S - F = \emptyset$},
\[
d(F, S) \: = \: {i \over k} \: < \: 1 \: \leq \: \mid S' \mid 
\: \leq \: \mid S' \mid \: + \: {j \over k}.
\]
Equation~\ref{eq:onethree} is satisfied.

In the sequel we shall assume, sometimes without recalling this explicitly, 
that the distance $d$ satisfies 
Equation~\ref{eq:onethree}.

Our main result is that stereotypical reasoning yielded by a distance
that satisfies the four assumptions above: 
i.e., uniqueness of the closest stereotype,
antimonotonicity of the distance \mbox{$d(F , S)$} in \mbox{$F \cap S$}, 
monotonicity in \mbox{$S - F$} and independence from \mbox{$F - S$},
is cumulative~\cite{KLMAI:89}.
The main result is therefore the following.
\begin{theorem}
\label{the:main}
If \mbox{$F \cap S^{F} \subseteq F' \subseteq F$}, then
\mbox{$S^{F'} = S^{F}$}.
\end{theorem}
\proof
Assume \mbox{$F \cap S^{F} \subseteq F' \subseteq F$}. We must show
that, for any stereotype $S$, we have
\mbox{$d(F' , S^{F}) \leq d(F' , S)$}.
First, since \mbox{$F \cap S^{F} \subseteq F'$},
we have both \mbox{$F \cap S^{F} \subseteq F' \cap S^{F}$}
and 
\mbox{$S^{F} - F' \subseteq S^{F} - F$}, therefore, by
Equation~\ref{eq:onethree}, we have
\mbox{$d(F', S^{F}) \leq d(F, S^{F})$}.
By Equation~\ref{eq:close}, for any stereotype $S$,
\mbox{$d(F, S^{F}) \leq d(F, S)$} and therefore, for any $S$,
\mbox{$d(F', S^{F}) \leq d(F, S)$}.
Using, now, \mbox{$F' \subseteq F$}, we see that
\mbox{$F' \cap S \subseteq F \cap S$} and 
\mbox{$S - F \subseteq S - F'$}. By Equation~\ref{eq:onethree},
then, \mbox{$d(F, S) \leq d(F', S)$},
and  \mbox{$d(F' , S^{F}) \leq d(F' , S)$}, for any stereotype $S$.
\QED
\begin{corollary}
The nonmonotonic consequence relation \NI\ defined by stereotypical
reasoning yielded by a distance
that satisfies Equation~\ref{eq:onethree} satisfies Cut and Cautious
Monotonicity and is therefore cumulative.
\end{corollary}
\proof
Suppose \mbox{$\alpha$ \NI $\beta$}.
Let $F$ be the set of models of $\alpha$ and $F'$ be the set of
models of \mbox{$\alpha \wedge \beta$}.
The assumption \mbox{$\alpha$ \NI $\beta$} means that all elements of
\mbox{$F \cap S^{F}$} satisfy $\beta$, i.e., 
\mbox{$F \cap S^{F} \subseteq F'$}.
But clearly \mbox{$F' \subseteq F$}.
By Theorem~\ref{the:main}, \mbox{$S^{F'} = S^{F}$}
and \mbox{$ F \cap S^{F} = F' \cap S^{F'}$} and
\mbox{$\alpha$ \NI $\gamma$} iff \mbox{$\alpha \wedge \beta$ \NI $\gamma$}. 
\QED
Karl Schlechta~\cite{Schl:counter} has found a cumulative consequence 
relation \NI\ that cannot
be defined by any stereotypical reasoning system yielded by a distance
that satisfies Equation~\ref{eq:onethree}.
The exact characterization of those cumulative relations that can be
defined by stereotypical systems that satisfy Equation~\ref{eq:onethree}
is open.
In the next section, we shall discuss another basic logical property
of nonmonotonic system described in~\cite{KLMAI:89}, Or, i.e., 
preferentiality.
\section{Preferentiality}
Suppose each one of two information sets, $F$ and $F'$ enable us
to conclude that some formula $\alpha$ holds:
all elements of \mbox{$F \cap S^{F}$} and all elements of
\mbox{$F' \cap S^{F'}$} satisfy $\alpha$.
Does this imply that the union \mbox{$F \cup F'$} enables us
to conclude $\alpha$, i.e., do all elements of
\mbox{$(F \cup F') \cap S^{F \cup F'}$} satisfy $\alpha$?
The discussion of~\cite{KLMAI:89} explains why this seems to be a natural
property to expect.
For example, assuming we would conclude that a bird that lives in the country
flies, and that we would also conclude that a bird that lives in a city
flies.
Must we conclude that birds that live either in the country
or in a city fly?
Stereotypical reasoning does not always satisfy
this property for the following reason.
I guess that natural common-sense reasoning does not either, 
for the same reason.
Suppose $\beta$ and $\gamma$ describe very different situations, whose best 
stereotypes are different. It may happen, nevertheless that the same
property $\alpha$ will be shared both by models 
of \mbox{$\beta \cap S^{\beta}$} and of \mbox{$\gamma \cap S^{\gamma}$}.
But the best stereotype for \mbox{$\beta \vee \gamma$} may be very general
and some models of, say, \mbox{$\beta \cap S^{\beta \vee \gamma}$}
may not satisfy $\alpha$.
Intuitively, if the reasons for concluding $\alpha$ from $\beta$
are very different from those for concluding $\alpha$ from $\gamma$,
there is little hope we shall be able to conclude $\alpha$ from the
disjunction \mbox{$\beta \vee \gamma$}.

There is one interesting case, though, the case \mbox{$S^{F} = S^{F'}$},
in which the desired conclusion follows if we strengthen one of the 
assumptions above. Since the function $e$ does not depend 
on its third argument, by Assumption three, we shall write it
as a function of two arguments.
Let us assume:
\begin{itemize}
\item {\em Assumption four}: 
\mbox{$e(A \cup A' , B) = \min\{e(A , B) , e(A' , B)\}$}. 
\end{itemize}
Clearly, assumption one already implies
\mbox{$e(A \cup A' , B) \leq \min\{e(A , B) , e(A' , B)\}$} and
assumption four implies assumption one.
\begin{theorem}
\label{the:or}
Let assumptions zero--four be satisfied.
If \mbox{$S^{F} = S^{F'}$}, 
then \mbox{$S^{F \cup F'} = S^{F}$}.
\end{theorem}
\proof
Notice that we do not claim that the nonmonotonic consequence
relation defined is preferential.
Assume \mbox{$S^{F} = S^{F'}$}.
We must show that, for any stereotype $S$, we have
\mbox{$d(F \cup F' , S^{F}) \leq d(F \cup F' , S)$}.
Then, 
\[
d(F \cup F' , S_{F}) = 
e((F \cap S_{F}) \cup (F' \cap S_{F'}) , 
S_{F} - (F \cup F') ) \leq
\]
\[
e(F \cap S_{F} , S_{F} - F) = 
d(F , S_{F}) \leq
d(F , S). 
\]
Similarly \mbox{$d(F \cup F' , S_{F}) \leq d(F' , S)$} and
\[
d(F \cup F' , S_{F}) \leq \min\{d(F , S) , d(F' , S)\} = d(F \cup F' , S).
\]
The last equality stems from assumption four.
\QED
Consider our examples above.
In Example~\ref{ex:1}, the function $d$ is constant and therefore satisfies
condition four. The consequence relation, being classical, is in fact
preferential.
In Example~\ref{ex:2}, the function $d$ proposed does not satisfy
assumption four, nevertheless the relation defined is preferential.
In Example~\ref{ex:3}, the function $d$ satisfies assumption four. 
The consequence relation defined, being classical, is in fact preferential.
In Example~\ref{ex:4}, assumption four holds, and therefore the conclusions
of Theorem~\ref{the:or}, but the consequence relation defined is {\em not}
preferential.
\section{Conclusion}
A formal description of stereotypical reasoning has been provided.
Under reasonable assumptions about the way stereotypes are attached
to information sets, this model yields a cumulative system.
The assumptions one--three concerning the distance between information
sets and stereotypes may perhaps be tested experimentally.
Preferentiality has been discussed, and found unplausible in general,
but a more limited natural property has been put in evidence.
Again, preferentiality should be tested experimentally.
The conditions proposed above that imply good logical behavior
are sufficient but not necessary.
Other conditions may be more natural and also sufficient.
The structure of the set $S$ of stereotypes, in particular,
has been left completely arbitrary.
A reasonable assumption may be that this set has a tree structure:
i.e., that if $S$ and $T$ are any two stereotypes such that
the intersection \mbox{$S \cap T$} is not empty, then 
\mbox{$S \subseteq T$} or \mbox{$T \subseteq S$}. 
\section{Acknowledgments}
I would like to thank the members of the different audiences
that reacted to the material above, while it was in gestation,
and in particular to Yuri Gurevich, Pierre Livet, Drew McDermott,
Karl Schlechta and Moshe Vardi.
This work was partially supported by the Jean and Helene Alfassa fund for 
research in Artificial Intelligence and by grant 136/94-1 of the 
Israel Science Foundation on ``New Perspectives on Nonmonotonic Reasoning''.

\bibliographystyle{plain}

\end{document}